%% file: main.tex
\documentclass[letterpaper, 10 pt, conference]{ieeeconf}
\IEEEoverridecommandlockouts 
\usepackage{graphics,epsfig,epstopdf,mathptmx,times,amsmath,amssymb,psfrag,mathtools,graphicx}
\usepackage{cite}
\usepackage[utf8]{inputenc}
\usepackage[english]{babel}
\usepackage{mleftright}
\usepackage[makeroom]{cancel}
\usepackage{tikz}
\usepackage{siunitx}
\usepackage{balance}
\usepackage{subcaption}
\usepackage{flushend}
\usepackage{floatrow}

\title{\LARGE \bf
Insights from an Industrial Collaborative Assembly Project: \\ Lessons in Research and Collaboration}
\author{Tan Chen$^{1}$, Zhe Huang$^{2}$, James Motes$^{3}$, Junyi Geng$^{4}$, Quang Minh Ta$^{1}$, Holly Dinkel$^{5}$, \\  Hameed Abdul-Rashid$^{3}$, Jessica Myers$^{3}$, Ye-Ji Mun$^{2}$, Wei-che Lin$^{6}$, Yuan-yung Huang$^{6}$, Sizhe Liu$^{6}$, \\ Marco Morales$^{3}$, Nancy M. Amato$^{3}$, Katherine Driggs-Campbell$^{2}$, and Timothy Bretl$^{5}$
\thanks{$^{1}$T. Chen and Q. Ta are with the Coordinated Science Laboratory at the University of Illinois at Urbana-Champaign, \{tanchen, minh\}@illinois.edu.}
\thanks{$^{2}$Z. Huang, Y. Mun, and K. Driggs-Campbell are with the Department of  Electrical and Computer Engineering at the University of Illinois at Urbana-Champaign, \{zheh4, yejimun2, krdc\}@illinois.edu.}
\thanks{$^{3}$J. Motes, J. Myers, H. Abdul-Rashid, M. Morales, and N. Amato are with the Department of Computer Science at the University of Illinois at Urbana-Champaign, \{jmotes2, jmmyers3, hameeda2, moralesa, namato\}@illinois.edu.}
\thanks{$^{4}$J. Geng is with the Robotics Institute at Carnegie Mellon University, junyigen@andrew.cmu.edu.}
\thanks{$^{5}$H. Dinkel and T. Bretl are with the Department of Aerospace Engineering at the University of Illinois at Urbana-Champaign, \{hdinkel2, tbretl\}@illinois.edu.}
\thanks{$^{6}$W. Lin, Y. Huang, and S. Liu are with Foxconn Interconnect Technology, \{wei-che.lin, yuan-yung.huang, sizhe.liu\}@fit-foxconn.com.}
}

\begin{document}
\maketitle
\thispagestyle{empty}
\pagestyle{empty}

\input{abstract}
\input{intro}
\input{description}
\input{lessons}
\input{conclusions}

\section*{Acknowledgment}
\noindent \small This work was performed with support from Foxconn Interconnect Technology (FIT) and the Center for Networked Intelligent Components and Environments (C-NICE) at the University of Illinois at Urbana-Champaign.

\bibliographystyle{IEEEtran}
\bibliography{IEEEabrv,bibfile}
\end{document}

%% file: abstract.tex
\begin{abstract}

Significant progress in robotics reveals new opportunities to advance manufacturing. Next-generation industrial automation will require both integration of distinct robotic technologies and their application to challenging industrial environments. This paper presents lessons from a collaborative assembly project between three academic research groups and an industry partner. The goal of the project is to develop a flexible, safe, and productive manufacturing cell for sub-centimeter precision assembly. Solving this problem in a high-mix, low-volume production line motivates multiple research thrusts in robotics. This work identifies new directions in collaborative robotics for industrial applications and offers insight toward strengthening collaborations between institutions in academia and industry on the development of new technologies.
\end{abstract}

%% file: intro.tex
\section{Introduction}
\label{introduction}

Challenges in automating manufacturing are important drivers for innovation in disciplines like computer vision, state estimation, motion planning and control, and machine learning. Picking applications requiring knowledge of object state inspired the development of new 6D pose estimation from RGB imagery using traditional computer vision methods \cite{collet2011moped,hinterstoisser2011gradient} and machine learning-based methods \cite{brachmann2014learning,krull2015learning,deng2021poserbpf}.
Assembly has also made advancements, from model-based methods \cite{stemmer2007analytical, liu2014high} to learning-based methods \cite{inoue2017deep,luo2019reinforcement}, for high-precision assembly operations in industry. Progress in human-robot interaction optimizes across the skills of both people and robots by studying their safe and efficient collaboration on various industrial tasks \cite{cheng2020towards, wilbert2012robot, shi2012levels, grahn2016potential, tsarouchi2017human}. Recent advancements in multi-robot task and motion planning account for precedence constraint assembly tasks \cite{brown2020optimal}, leverage multi-arm cooperation \cite{shome2020synchronized,chen2022cooperative}, and find collision-free trajectories for large teams of robots \cite{solis2021representation}.

This paper reports lessons learned from a collaborative assembly project working with an industry partner. The collaborative assembly project aims to integrate modern technologies in robotics \emph{research} and develop an advanced manufacturing cell for high-precision assembly to boost productivity in \emph{industry}. Research challenges primarily arise from environmental constraints, namely the contact-rich environment inherent in the assembly task, and task requirements such as autonomy and flexibility. Moreover, this paper discusses the challenges in university-industry collaboration that arise due to their distinct missions and cultures. Awareness of these challenges suggests new directions for industrial robotics and promotes efficiency in university-industry collaborations.

%% file: description.tex
\section{Collaborative Industrial Assembly}
\label{sec:description}

\begin{figure}
    \centering
    \includegraphics[width=0.92\columnwidth]{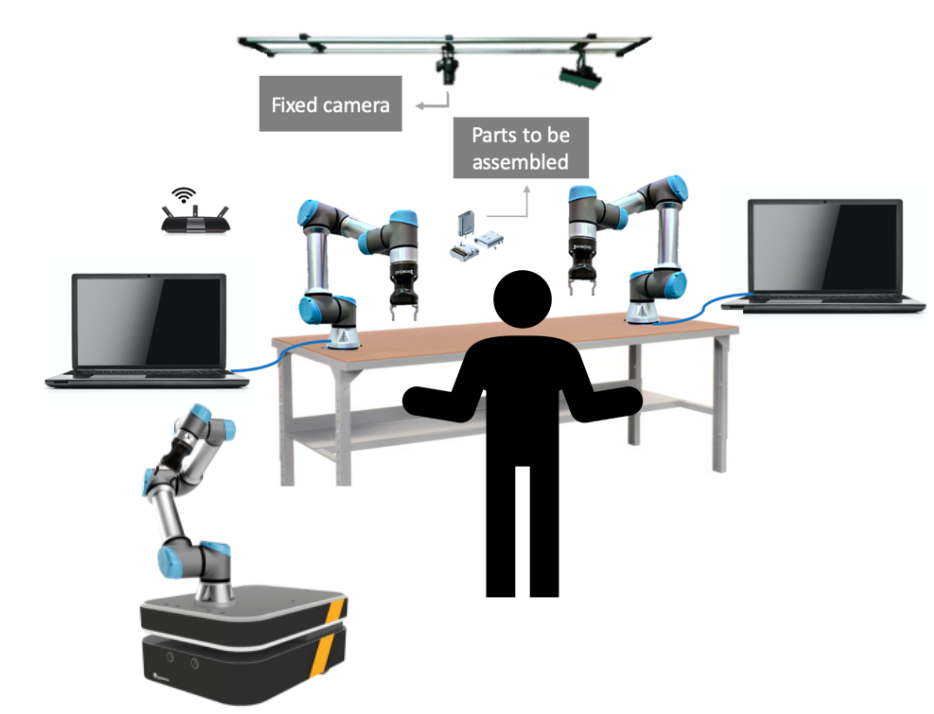}
    \caption{An efficient industrial assembly cell integrates perception, control, planning, and scheduling in a contact-rich environments while considering safety and flexibility.}
    \label{fig:cell}
\end{figure}

The collaborative assembly cell for this project is shown in Figure \ref{fig:cell}. Two UR5e robotic manipulators each with Robotiq Hand-E parallel grippers are fixed on a table to assemble small parts \cite{ur5e, hande}. The part localization and pose information is computed from RGB images provided by a Basler ace camera mounted overhead \cite{basler}. A UR5e affixed to a Clearpath Boxer mobile robot acts as a mobile manipulator \cite{clearpath}, fetching new parts for assembly and transporting assembled components to successive assembly lines. The robots, camera, and computers communicate perception information and plans with each other via the Robot Operating System (ROS) message-passing middleware \cite{ros}. Operators can assist the robots with assembly in the shared workspace. This project designs, develops, and tests a \emph{flexible}, \emph{safe} and \emph{productive} manufacturing cell for \emph{high-precision} assembly, where these requirements are defined as follows:

\begin{itemize}
    \item \textbf{Flexibility:} The cell is designed for application in high-mix, low-volume production lines, with quick adaptation of the production cell for assembly of new parts.
    \item \textbf{Safety:} Robotic and human agents within the manufacturing cell have knowledge about the states of other agents and avoid collisions in task execution.
    \item \textbf{Productivity:} The manufacturing cell reduces the number of operators in low-volume production lines by 75\% and increases the production rate by 50\%.
\end{itemize}

This goal motivates \emph{Picking}, \emph{Assembly}, \emph{Safety}, and \emph{Scheduling} research thrusts. The focus of the Picking research thrust is on solving the problem of reliably picking parts from a cluttered bin -- a task on which everything else depends; the focus of the Assembly research thrust is on solving the problem of joining parts prior to pressing or welding; the focus of the Safety research thrust is on solving the problem of human-robot interaction to enable safe operation in the shared workspace and part transfer between human and robot workers; and the focus of the Scheduling research thrust is on solving the problem of allocating tasks among multiple robots and of planning fast, collision-free trajectories for these robots.

%% file: lessons.tex
\section{Industrial Assembly Insights}
\label{sec:lessons}
There are two salient challenges in developing a collaborative assembly cell for industrial use: the challenges involved with research and development of new technologies and the challenges involved with collaboration between university and industry. We summarize the challenges in the research aspect which arise mostly from environmental constraints and task specifications. We also provide potential directions for effective integration of the collaborative assembly cell.

\subsection{Contact-Rich Environments}
\label{subsec:contact}
\begin{figure}
     \centering
     \begin{subfigure}{0.46\textwidth}
         \centering
         \includegraphics[width=\textwidth]{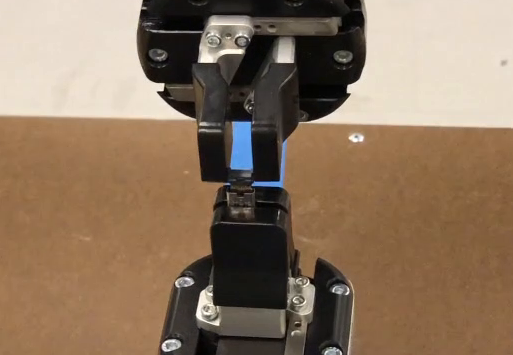}
         \caption{insertion}
         \label{fig:insert}
     \end{subfigure}
     \begin{subfigure}{0.46\textwidth}
         \centering
         \includegraphics[width=\textwidth]{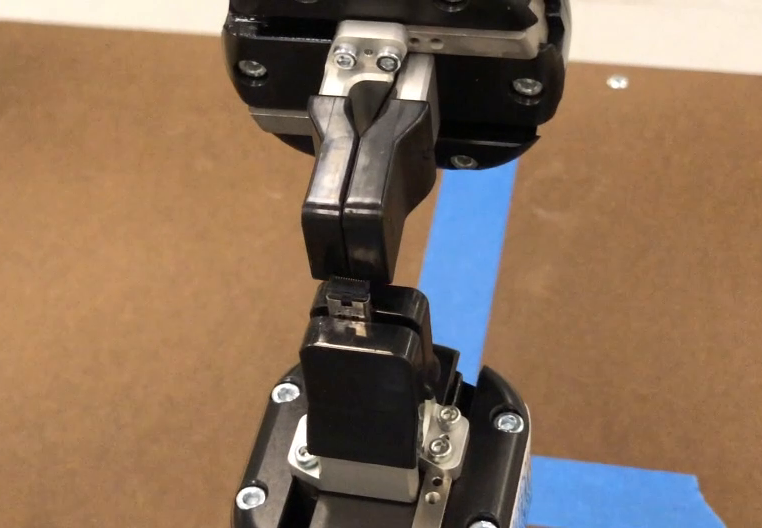}
         \caption{push from side}
         \label{fig:sidepush}
     \end{subfigure} \\
     \begin{subfigure}{0.46\textwidth}
         \centering
         \includegraphics[width=\textwidth]{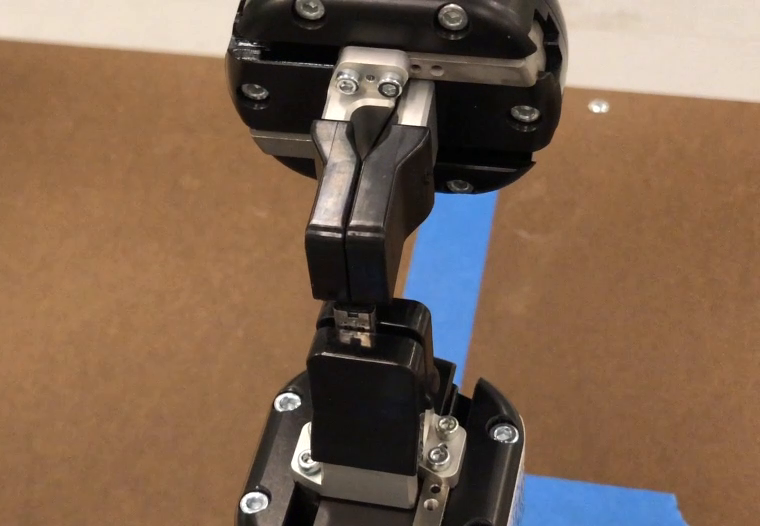}
         \caption{push from top}
         \label{fig:toppush}
     \end{subfigure}
     \begin{subfigure}{0.46\textwidth}
         \centering
         \includegraphics[width=\textwidth]{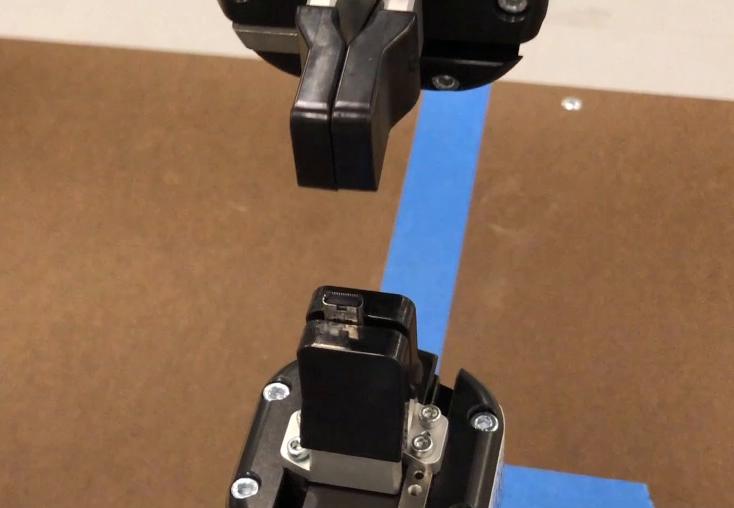}
         \caption{assembly completed}
         \label{fig:complete}
     \end{subfigure}
    \caption{A complete assembly process, specifically for assembling two parts (an insert part and a case part) in this project, requires insertion, followed by two pushes (``push-push'').}
    \label{fig:assembly}
\end{figure}
The collaborative assembly cell is a contact-rich environment -- there are contacts among human, robots, parts, and the environment. 
Figure~\ref{fig:assembly} shows several key steps to accomplish a complete assembly of an insert part and a case part: the bottom gripper is fixed holding the case part and the top gripper moves the insert part to complete assembly. 
During the insertion stage in Figure~\ref{fig:insert}, note that there is contact interference between the robot finger and the case part. 
Therefore, the assembly cannot be completed by only one insertion step which is the strategy of most peg-in-hole problems \cite{stemmer2007analytical,liu2014high,inoue2017deep,luo2019reinforcement}. 
The assembly for this industrial project requires additional steps to tune the assembly, for example, a push-push strategy as shown in Figures~\ref{fig:sidepush} and \ref{fig:toppush} to ensure the insert part is well-inserted into the case part.

Because contact results in discontinuous motion and uncertain states after contact, it causes more difficulty in planning. 
Taking the aforementioned assembly as an example, there is no continuous motion to complete the assembly. A push-push strategy is adopted, despite the fact that the strategy is engineered. 
An advanced collaborative assembly cell should be able to reverse engineering and automate such a decision. 
In a broader sense, one major challenge for a collaborative assembly cell is generating a feasible and robust task and motion plans in contact-rich environments. 
One possible direction to tackle this challenge is designing action items (such as push-push strategy) that can be added in a library for the planner.

\begin{figure}
    \centering
    \includegraphics[width=\columnwidth]{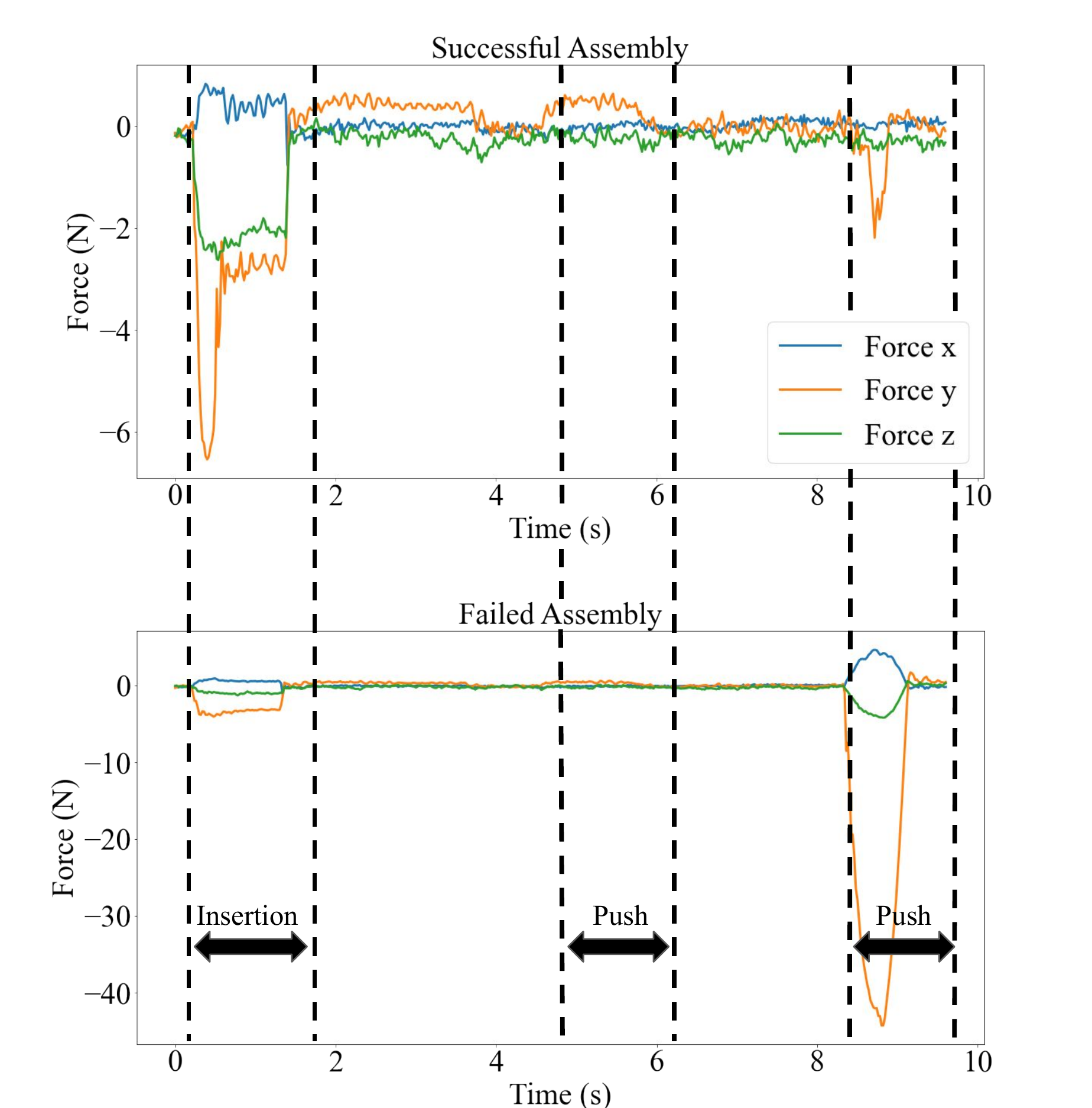}
    \caption{A comparison of force detected by a F/T sensor at assembly. Top plot is for a successful assembly, and bottom plot is for a failed assembly. The three key steps, insertion, push from side and push from top, are also marked. Also note that they have separate scales along y-axes.}
    \label{fig:contact_cp}
\end{figure}

Contact can also be leveraged in positive ways. The low sampling rate of a camera and occlusion limit the ability of using visual systems for online tuning in the assembly. In contrast, the force torque (F/T) sensor at the robotic wrist can detect contact forces during assembly and have a much higher sampling frequency. Thus, some important problems along this direction are exploiting the contact information to improve the assembly. For example, Figure~\ref{fig:contact_cp} shows a comparison of contact force (detected by a F/T sensor) between a successful and failed assembly, which demonstrates that the contact information contains rich features to distinguish between successful and failed assemblies. Due to data noise and model complexity, applying learning-based methods to analyze the contact information is promising.

The environment can also be designed to leverage contact. A common issue that we have observed in experiments is the sticky gripper, which causes difficulty in dropping the assembled parts. 
An effective solution is building a collection bin and leveraging the contact of a slight nudge between the part and edge of the bin for dropping. 
This suggests a research direction in building an optimized environment to minimize the effect of engineering issues and maximize the success rate of assembly.

\subsection{Uncertainty}
Precision assembly relies on highly precise pose estimation, which can suffer from noises and errors at each level of the \emph{perception pipeline}, such as camera intrinsic and extrinsic calibration, raw image processing, and pose estimation algorithm. As many pieces are metal, reflective, and tiny, infrared depth sensors from RGBD cameras such as Intel RealSense are not suitable for distance measurement. Pose estimation has to use RGB images. The appearance of the metal pieces depends on lighting conditions, which affect the performance of pose estimation. Thus, it is required to either have a robust pose estimation algorithm that can handle various lighting conditions, or the lighting condition has to be strictly controlled in a factory to match the setup during training. Occlusion due to cluttered pieces in a bin is also a challenge for reliable pose estimation.

Part pose uncertainty also comes from the assembly process. A slight part misalignment can lead to an unexpected contact point between the pieces. Since the misalignment is typically difficult to be detected via vision, the assembly process will continue, and the contact may yield a torque which can result in slip between the in-hand piece and the gripper, and a failed assembly. Pose estimation 
needs to address the situation where the piece is occluded by the gripper. Otherwise, the robot must deal with the uncertainty during contact interactions.

An alternative direction to handle part pose uncertainty is to introduce human-robot collaboration, where human cognition can be well integrated into robot automation. However, humans are a source of uncertainty themselves. Human body pose and intention need to be effectively detected, estimated, tracked and even predicted to achieve safe and efficient human-robot collaboration, and uncertainty exists at each level of the \emph{collaboration pipeline}. While motion capture offers decent human body pose tracking~\cite{lasota2014toward}, wearable sensors can potentially constrain the flexibility of human dexterous manipulation on the small pieces, and the tracking system is prohibitively expensive when deployment is scaled up. A vision-based human body detection algorithm OpenPose~\cite{cao2019openpose} and Kalman filter are adopted for human body pose tracking. This solution is cheap, but the tracking performance can be easily degraded due to the frequent occlusion when human and robot are in close proximity. Cameras from multiple views are used to minimize the influence from the occlusion. To enforce safety of human and robot, the robot receives not only vision signals such as whether human body is detected within the robot work space, but also haptic signals such as whether an unexpected contact happens to perform safety-rated monitored stop.

\begin{figure}[t]
    \centering
    \includegraphics[width=\linewidth]{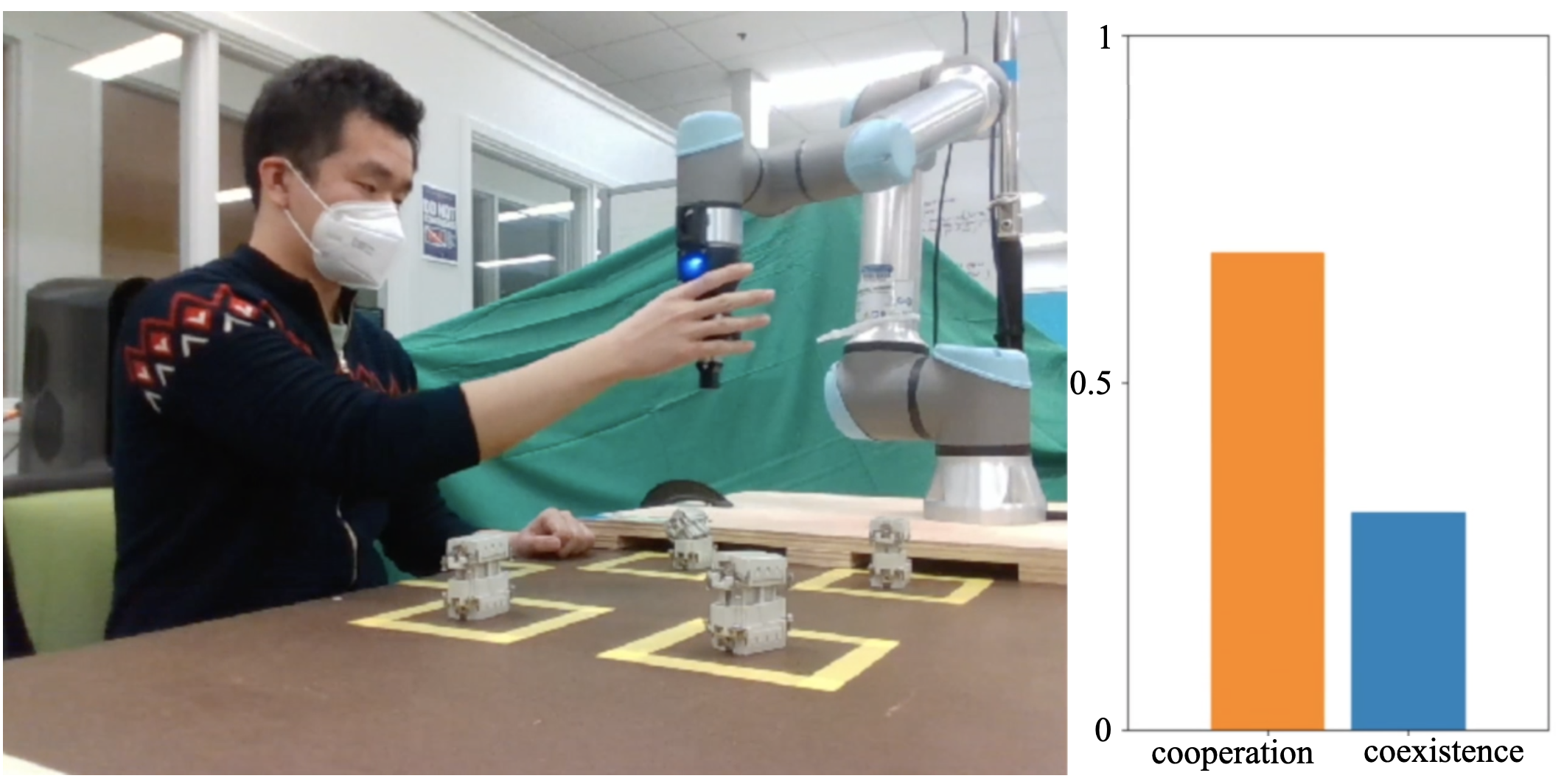}
    \caption{A use case of collaborative assembly. Human needs to align parts, and robot needs to push aligned parts together. Different interaction patterns are present in this task. Coexistence intention means human wants concurrent task execution. Cooperation intention means human wants to manually guide the robot. The robot not only needs to track which part pair the human is working on, but also estimate interactive intention to switch to the corresponding interactive mode for seamless collaboration. The bar plot shows the probability distribution of human interactive intention estimate.}
    \label{fig:use-case}
    \vspace{-10pt}
\end{figure}

Human motion is generally intention-driven in collaborative assembly tasks. Intention tracking and motion prediction should be concurrently implemented and can benefit each other~\cite{huang2021long}. Figure~\ref{fig:use-case} illustrates a developed use case, where a human aligns male and female parts, and a robot pushes aligned pairs to finish assembly~\cite{huang2022hierarchical}. Taking tracked human motion as input, low-level task intention tracking identifies which pair the human is working on, and robot plans corresponding motion to seamlessly collaborate with the human. Taking both human and robot motion as input, high-level interactive intention tracking estimates whether the human desires to focus on part alignment without interruption from robot (coexistence intention), or guide the robot to recover previously failed pushing attempt (cooperation intention). Though promising intention tracking performance is demonstrated in our work~\cite{huang2022hierarchical}, incorrect intention estimates could still happen when observations (for example tracked human positions) are noisy or even degraded, and when novel unseen human behavior is present. The robot control architecture should be robust to uncertainty in intention, as the robot has to safely recover to the original state from the plan based on a wrong intention estimate. In addition, artificial potential field, which we used for motion planning, provides reactive control by treating the human body as dynamic obstacles. We believe effective integration of human motion prediction into robot motion planning will provide proactive control that avoids local optima and freezing robot problems, and further improves collaboration efficiency~\cite{unhelkar2018human}.

\subsection{Versatility}

\begin{figure}
\floatbox[{\capbeside\thisfloatsetup{capbesideposition={right,top},capbesidewidth=4cm}}]{figure}[\FBwidth]
{\caption{A notch is ``crafted'' at the finger tip to fit the shape of the part for accurate localization of the part.} \label{fig:fingers}}
{\includegraphics[width=0.3\textwidth]{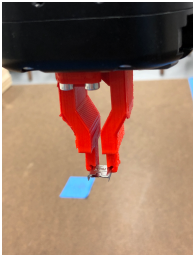}}
\end{figure}

The collaborative assembly cell aims to assemble different parts. 
Because of the small size (in \si{mm} level), irregular shape, and fragility of the parts, \emph{grasping is a big challenge caused by versatility}. 
The two-finger parallel gripper adopted in the project is versatile in most cases, but it is too rigid to complete the assembly with only one insertion (as discussed in Section \ref{subsec:contact}). 
Inspired by human operators working on the assembly, soft robotics is a promising direction to develop a versatile and soft gripper for grasping. 
The other approach that we are exploring is designing a finger specially ``crafted'' at the tip to best fit the shape of parts and passively guide the part into the notch at grasping (see Figure~\ref{fig:fingers}). 
A major advantage of the design is that it can indirectly provide an accurate relative position between the part and the robot base despite some uncertainty from pose estimation. 

Besides grasping, other research thrusts in the project must also consider versatility. 
For example, the pose estimation pipeline should be able to work on different part types. 
Thus, a large dataset including the various part types is required to train a model for pose estimation. 
Collection of real image datasets is time-consuming and error-prone. Synthetic dataset generation is a favorable alternative to collect the data for training part detection and state estimation models \cite{dilmegani}. 
Control and planning algorithms also need to be versatile to handle different part types. 
As one major goal of the project is to enable quick change and update of the production line for assembling new parts, machine learning methods, such as transfer learning, should be useful in tuning parameters.

\subsection{University-Industry Collaboration}
There are many benefits from collaboration between university and industry such as driving innovation and positively impacting society \cite{venturewell}. 
However, differences in philosophies between university and industry pose challenges to the collaboration. 
In contrast with academic research, which explores state-of-the-art methods and seeks to develop novel technological contributions, industry expects an applicable and complete solution with immediate economic benefits. 
This results in a gap between the goals of the project for the respective parties. The strengths of university and industry are also different. At a university, the focus is on prototyping and testing novel techniques. The principal investigators leading different research thrusts in a large industrial project focus on their fields of expertise and may find it challenging to efficiently integrate and mature a product. Furthermore, industry partners typically have more resources and experience in product development but may be less familiar with the research and development life cycle. 

Additionally, project management styles in academia and industry are different. 
Because of the exploratory nature of research, academia tends to adopt more agile methodologies, which provide flexibility but also require more self-motivation and interaction. Project management in industry generally adopts a rigid waterfall structure to ensure productivity \cite{leeronh}. However, university-industry collaborated projects need an interface to translate the research progress to the stakeholders in the company. A management approach balancing the agile and waterfall methodologies should be more suitable for such projects. For example, the industry and university can make a plan together (waterfall) according to the statement of work of the project. In the project execution, while the team tries to push the project forward on time, some flexibility should be allowed to discuss the feasibility of individual project items and allow alternative plans during weekly meetings (agile).

To set win-win project goals and make feasible schedules, being aware of the differences and a deep dive into relevant domain knowledge are necessary for both parties to understand the prominent issues in industry and how the outcomes of academic research can improve the production in industry.

%% file: conclusions.tex
\section{Conclusions}
\label{sec:conclusions}

This paper reports several lessons learned from a collaborative assembly project working with an industry partner. The challenges of the project have been discussed in terms of multiple aspects related to research and university-industry collaboration. Based on these challenges, several research directions are identified and shared for effective integration of collaborative robotics for industrial application. Being aware of the differences between university and industry and developing a suitable work-plan are also the key to successful transfer of academic research to industrial application.